\pdfoutput=1
\documentclass[11pt]{article}

\usepackage{EMNLP2022}

\usepackage{times}
\usepackage{latexsym}
\usepackage{todonotes}
\usepackage{multirow}
\usepackage{multicol}
\usepackage{lingmacros}
\usepackage{float}
\usepackage{dblfloatfix}
\usepackage{hyperref} 

\usepackage[T1]{fontenc}
\usepackage[utf8]{inputenc}

\usepackage{microtype}

\usepackage{inconsolata}

\title{Leveraging a New Spanish Corpus for Multilingual and Crosslingual Metaphor Detection}

\author{Elisa Sanchez-Bayona \and Rodrigo Agerri \\
  HiTZ Center - Ixa, University of the Basque Country UPV/EHU \\
  \texttt{elisa.sanchez@ehu.eus}, \texttt{rodrigo.agerri@ehu.eus} 
  \\}

\begin{document}
\maketitle
\begin{abstract}
The lack of wide coverage datasets annotated with everyday metaphorical expressions for languages other than English is striking. This means that most research on supervised metaphor detection has been published only for that language. In order to address this issue, this work presents the first corpus annotated with naturally occurring metaphors in Spanish large enough to develop systems to perform metaphor detection. The presented dataset, CoMeta, includes texts from various domains, namely, news, political discourse, Wikipedia and reviews. In order to label CoMeta, we apply the MIPVU method, the guidelines most commonly used to systematically annotate metaphor on real data. We use our newly created dataset to provide competitive baselines by fine-tuning several multilingual and monolingual state-of-the-art large language models. Furthermore, by leveraging the existing VUAM English data in addition to CoMeta, we present the, to the best of our knowledge, first cross-lingual experiments on supervised metaphor detection. Finally, we perform a detailed error analysis that explores the seemingly high transfer of everyday metaphor across these two languages and datasets.
\end{abstract}

\section{Introduction}

Metaphor can broadly be defined as the interpretation of a concept belonging to one domain in terms of another concept from a different domain \citep{Lakoff80metaphorswe}. Metaphorical expressions are recurrent in natural language as a mechanism to convey abstract ideas through specific experiences related to the real, physical world or to send a stronger message in a discourse. There is a large body of work from various fields such as linguistics, psychology or philosophy that tried to provide a theoretical characterization of metaphor. Some approaches are based on the semantic similarity shared between the domains involved \citep{gentner1983structure, kirby1997aristotle}, while others explain metaphorical uses of language in terms of violation of \emph{selectional preferences} \citep{wilks1975preferential, wilks1978making}. Other perspectives focus on the communicative impact of using a metaphorical expression in contrast to its literal counterpart \citep{searle1979expression, black1962models}. Following previous work on metaphor detection in Natural Language Processing (NLP) \citep{steenetal2010, reportleong-etal-2018-report}, our approach is based on the Conceptual Metaphor Theory of \citet{Lakoff80metaphorswe}. They do not conceive metaphors just as a cognitive-linguistic phenomenon commonly used in our everyday utterances. Instead, metaphors are understood as a conceptual mapping that typically reshapes an entire abstract domain of experience (target) in terms of a different concrete domain (source).

The high frequency of metaphors in everyday language has increased the popularity of research for this type of figurative language in the NLP field. One of the reasons behind is the fact that the automatic processing of metaphors is essential to achieve a successful interaction between humans and machines. In this sense, it is considered that other NLP tasks performance could benefit from metaphor processing, such as Machine Translation \citep{mao-etal-2018-word}, Sentiment Analysis \citep{Zhang2010MetaphorIA, rentoumi-etal-2009-sentiment}, Textual Entailment \citep{agerri2008metaphor, neubig} or Hate Speech Detection \citep{hatespeech, lemmens-etal-2021-improving}, among others.

However, the large majority of research on metaphor detection has been done for English, for which the public release of the VUAM dataset within the FigLang shared tasks from 2018 and 2020 marked a major milestone \cite{reportleong-etal-2018-report,leong-etal-2020-report}. In this paper we would like to contribute to research in multilingual and cross-lingual metaphor detection by presenting a new wide coverage dataset in Spanish with annotations for everyday metaphorical expressions.

In this context, the contributions of this work are the following: (i) A new publicly available dataset for metaphor detection in Spanish from a variety of domains, CoMeta; (ii) an in-depth discussion of problematic cases and of adapting the MIPVU method to annotate metaphor in Spanish; (iii) a quantitative and qualitative analysis of the resulting CoMeta corpus; (iv) competitive baselines using 18 large monolingual and multilingual language models in monolingual and cross-lingual evaluation settings, showing that modern language models such as DeBERTa \citep{He2021DeBERTaV3ID} perform similarly to models specifically trained for metaphor processing like MelBERT \cite{melbert}; (v) error analysis shows that, for these languages and datasets, cross-lingual metaphor transfer is very high, mostly due to the presence of metaphorical usage of commonly used verbs; (vi) the CoMeta dataset, code and fine-tuned models are publicly  available \footnote{\url{https://ixa-ehu.github.io/cometa/}} to encourage research in multilingual and cross-lingual metaphor detection and to facilitate reproducibility of results.

\section{Previous Work}

Metaphorical expressions can be conveyed through multiple linguistic structures and can be classified according to different criteria \citep{survey}. A common distinction is that of \textbf{conventional}  metaphors (as in Example (\ref{ex:conventional})), highly extended among speakers and lexicalized, and \textbf{novel} metaphors (Example (\ref{ex:novel})), which are less frequent in everyday utterances (examples taken from \citet{survey}).

\enumsentence{\label{ex:conventional} \emph{Sweet} love.} 
\enumsentence{\label{ex:novel} Snow \emph{debuts} on Twitter. } 

\citet{Lakoff80metaphorswe} argue that metaphors express a mapping across a source and targe domain which constitute a \textbf{conceptual metaphor}. Conceptual metaphors can be expressed through language resulting in \textbf{linguistic metaphors}. These in turn can be classified as \textbf{lexical metaphors} (as in (\ref{ex:conventional}), (\ref{ex:novel}) and (\ref{ex:nominal})) \textbf{multi-word metaphors} (\ref{ex:wipe}), and \textbf{extended metaphors}, which cover longer fragments of speech. With respect to the grammatical category to which they belong, we can find \textbf{verbal} (\ref{ex:novel}), \textbf{adjectival} (\ref{ex:conventional}), \textbf{nominal} (\ref{ex:nominal}) or \textbf{adverbial} metaphors (\ref{ex:adv}). 

\enumsentence{\label{ex:nominal} My lawyer is an old \emph{shark}}
\enumsentence{\label{ex:adv} Ram speaks \emph{fluidly}.}
\enumsentence{\label{ex:wipe}If you use that strategy, he’ll \emph{wipe} you \emph{out}. \citep{Lakoff80metaphorswe}}

Automatic processing of metaphor is generally divided into three different tasks: \textbf{detection} of metaphorical expressions, their \textbf{interpretation}, namely, the identification of the literal meaning expressed by the linguistic metaphor, and the \textbf{generation} of new metaphorical expressions. From here on, we will center our attention on metaphor detection.

Most work on metaphor detection has focused on English texts. The VU Amsterdam Metaphor Corpus (VUAMC or VUAM) \citep{steenetal2010} is the most extensive dataset with annotations for the characterization of linguistic metaphor. It consists of English texts  labeled with several typologies of metaphor following the the VU Metaphor Identification Procedure (MIPVU), discussed in Section \ref{sec:mipvu}. It was subsequently adapted to other languages \citep{mipvumultilang}. However, Spanish was not included and, for the languages that were, this adaptation did not include the development of annotated corpora.

First attempts to tackle metaphor detection in English were corpus-based \citep{charteris2004corpus, skorczynska2006readership, semino2017corpus}. Most recent approaches address the task as sequence labeling usually based on deep learning, neural networks and word embeddings \citep{wu2018thu, bizzoni2018bigrams}. In addition, syntactic and semantic features (WordNet, FrameNet, VerbNet, dependency analysis, morphology, etc.) are exploited in order to boost the performance of such models. The celebration of the 2018 and 2020 shared tasks \citep{reportleong-etal-2018-report, leong-etal-2020-report} around the detection of metaphors using the VUAM dataset contributed to a huge jump in development and performance, although top results were achieved by classifying mostly conventional metaphors \citep{tong-etal-2021-recent, neidlein-etal-2020-analysis}.

Others combine metaphor theories as features in addition to annotated data to feed pre-trained models based on the Transformers architecture \citep{devlin2018bert}. For instance, the state-of-the-art system MelBERT \citep{melbert} uses the Metaphor Identification Procedure (MIP) \citep{pragglejazmip} and \emph{selectional preferences} \citep{wilks1975preferential, wilks1978making, percy1958metaphor}. These theories argue that terms with matching semantic features tend to appear in the same context, and metaphors usually do not comply with this hypothesis. Furthermore, the recently published model of MIss RoBERTa WiLDe \citep{robertawilde} benefits from dictionary definitions as an additional feature to train their model based on the architecture of MelBERT.

Due to the lack of labeled data to train supervised models, previous work addressing Spanish metaphor processing has mainly been based on unsupervised approaches. However, as it is often the case for many other NLP tasks, unsupervised approaches obtain far lower results than supervised methods \citep{tsvetkov2014metaphor, shutova2017multilingual}. Other issues are that most work in Spanish has focused either on a very specific type of conceptual metaphor \citep{williams2016get}, or on the characterization of metaphor in very domain-specific data \citep{martinez-santiago}. The  development of CoMeta aims to compensate this lack of resources for the Spanish language. To the best of our knowledge, it constitutes the largest dataset of general domain texts with metaphorical annotations in Spanish that, despite not reaching the size of the VUAMC corpus, can be used as a starting point, suitable to be extended and improved in the future, to further advance multilingual and cross-lingual methods for metaphor detection.

\section{Dataset Development}

In the following subsections we detail the creation process of our dataset CoMeta, including the data collection and annotation.

\subsection{Data Collection}

In order to compile a general domain dataset with natural language utterances and everyday language metaphors, we gathered samples from existing datasets of Spanish texts with linguistic annotations. As a result, CoMeta consists of 3633 sentences with metaphor annotations at token level from texts of multiple genres, such as blog, Wikipedia, news, fiction, reviews and political discourse, extracted from the following two sources. 

\noindent \textbf{Universal Dependencies (UD)}: We used the two largest Spanish treebanks annotated within the UD framework, which include linguistic information, such as Part Of Speech (POS), lemmas or dependencies: \textbf{AnCora}\footnote{\url{https://universaldependencies.org/treebanks/es_ancora/index.html}} and \textbf{GSD}\footnote{\url{https://universaldependencies.org/treebanks/es_gsd/index.html}}.
UD Spanish AnCora is a UD formatted version of the original Ancora Corpus \citep{ancora}. It contains 17680 sentences from the news domain, from which we randomly extracted 2000 sentences.
The GSD treebank is an automatic compilation of texts from miscellaneous domains, such as Wikipedia, blogs and reviews. We selected also randomly a subset of 1000 sentences out of 16013.
After preprocessing and filtering to remove duplicates, a total number of 2862 instances were finally included in CoMeta (1925 from Ancora, 937 from GSD and 771 from PD).

\noindent \textbf{Political Discourse (PD)}: In addition to UD texts, we manually collected political discourse transcripts from the Spanish \footnote{\url{https://www.lamoncloa.gob.es/consejodeministros/ruedas/Paginas/index.aspx}} and the Basque Government \citep{escribano-EtAl:2022:LREC}, five from each source. We chose this domain due to the higher frequency of appearance of metaphorical expressions, which are often used in order to convey a more powerful message \citep{political_disc, political}. From this source, we collected 771 sentences with automatic linguistic information added with UDPipe \citep{udpipe}.

\subsection{Annotation Process}

The labelling of CoMeta was mostly carried out by a single annotator, a Spanish native speaker and expert linguist over 3 months as part-time job. All annotations were revised up to a total of 6 times. Initial rounds consisted in annotating all kind of metaphorical expressions. Subsequent four rounds were dedicated to identify metaphorical expressions of each POS. Last two rounds were employed to revise annotations and resolve borderline cases. In order to evaluate the consistency of the annotations and inter-annotator agreement, 6 more Spanish linguists were also involved in the annotation of a subsample of the corpus. This procedure will be further described in next Subsection \ref{sec:agreement}. We decided to use binary labels following the approach of the VUAM versions used in the shared tasks recently mentioned.

\subsubsection{Annotation Guidelines} \label{sec:mipvu}

The task of metaphor annotation is inherently subjective, since it is sometimes based on personal experience and cultural knowledge. The Metaphor Identification Procedure (MIP) \citep{pragglejazmip} constituted an attempt to provide a systematic guideline that facilitates the process. It was later extended to MIPVU \citep{steenetal2010}, to cover ambiguous cases and address more thoroughly complex issues such as Multiword Expressions (MWE) or polysemy. The development of MIPVU resulted in the VUAM corpus \citep{steenetal2010}. This procedure was subsequently adapted to other languages in \citep{mipvumultilang}, although no wide coverage annotated corpus resulted from that adaptation. We followed the MIPVU guidelines to label CoMeta. In broad terms, it consists of the following steps:

\begin{enumerate}
    \item Read the entire text–discourse to establish a general understanding of the meaning.
    \item Determine the lexical units in the text–discourse.
    \item \begin{enumerate}
        \item For each lexical unit in the text, establish its meaning in context, that is, how it applies to an entity, relation, or attribute in the situation evoked by the text (contextual meaning). Take into account what comes before and after the lexical unit.
        \item For each lexical unit, determine if it has a more basic contemporary meaning in other contexts than the one in the given context. For our purposes, basic meanings tend to be:
                \begin{itemize}
                    \item More concrete; what they evoke is easier to imagine, see, hear, feel, smell, and taste.
                    \item Related to bodily action.
                    \item More precise (as opposed to vague).
                    \item Historically older.
                    Basic meanings are not necessarily the most frequent meanings of the lexical unit.

                \end{itemize}
        \item If the lexical unit has a more basic current–contemporary meaning in other contexts than the given context, decide whether the contextual meaning contrasts with the basic meaning but can be understood in comparison with it.
    \end{enumerate}
    \item If yes, mark the lexical unit as metaphorical.

\end{enumerate}

\begin{table*}[ht!] 
\begin{tabular}{ccc|cc|cc}
\multicolumn{1}{l}{} & \multicolumn{2}{c}{\textbf{CoMeta}} & \multicolumn{2}{c}{\textbf{UD}} & \multicolumn{2}{c}{\textbf{PD}} \\ \cline{2-7} 
 & \textbf{Met} & \textbf{No\_met} & \textbf{Met} & \textbf{No\_met} & \textbf{Met} & \textbf{No\_met} \\\cline{1-7} 
\multicolumn{1}{c}{\textbf{VERB}} & 873 & 9803 & 570 & 7560 & 303 & 2243 \\ 
\multicolumn{1}{c}{\textbf{NOUN + PROPN}} & 847 + 1 & 20118 + 8418 & 507+0 & 15790+7010 & 340+1 & 4328+1408 \\ 
\multicolumn{1}{c}{\textbf{ADJ}} & 396 & 6922 & 313 & 5413 & 83 & 1509 \\ 
\multicolumn{1}{c}{\textbf{ADV}} & 28 & 3836 & 15 & 2779 & 13 & 1057 \\ 
\multicolumn{1}{c}{\textbf{Total}} & 2145 & 49097 & 1405 & 38552 & 740 & 10545 \\ \hline
\end{tabular}
\caption{
Number of metaphorical and non-metaphorical tokens by POS in overall CoMeta and in the separate domains from Universal Dependencies (UD) and Political Discourse (PD).
}
\label{tab:data_analysis}
\end{table*}

\subsubsection{Scope of Annotations}

The definition of ``word'' provokes continuous and unsolved debates in the linguistics field. In MIPVU they use the more general term ``lexical unit'', understood as the basic piece that bears meaning, either a segment with its own POS or MWE. We followed this criterion as well in CoMeta. With regard to the POS, we decided to label only semantically significant classes: nouns, verbs, adjectives and adverbs, since most metaphors belong to one of these types. Details about the resulting dataset are reported in Table \ref{tab:data_analysis}.  

In this work, we focus on metaphorical expressions constrained to lexical units in the context of sentences. Thus, extended metaphors, where the figurative meanings are recurrent along larger pieces of texts, are not taken into account.

\subsubsection{Borderline Features}

\noindent \textbf{Other Forms of Figurative Language}: The boundaries between metaphor and other types of figurative language are not always clearly discernible. Specially in the case of metonymic expressions. 

In this work, we do not annotate metonymy, since we regard them as two different and distinguishable cognitive phenomena. In the case of metonymy, a concept is substituted by another from the same domain through a relationship of contiguity, e.g. \emph{beber una botella de ginebra (lit. ``to drink a bottle of gin'')}. In this example, the container is used to refer to the beverage but both terms belong to the domain of drink consumption. On the other hand, metaphorical expressions associate two different concepts from two distinct domains. With respect to similes, we treat them as a form of metaphor with a linguistic cue that makes the association of concepts explicit, e.g. ``like". Thus, similes are annotated in the same way as metaphors, marking the lexical units with figurative meaning. 

\noindent \textbf{Polysemy}: MIPVU’s guidelines establish a comparison between the contextual meaning
of a lexical unit and a more basic one in order to spot metaphors. However, some cases are ambiguous, due to polysemy, and can lead to confusion in the annotation process. For instance, in the example (\ref{ex:claro-reina}) from CoMeta, the adjective \emph{claro} (lit. ``clear'') presents various basic meanings in Diccionario de la Real Academia Española (DRAE) \citep{drae}: ``Que tiene abundante luz'' (lit. ``Having abundant light'') and ``Dicho de un color o de un tono: 
Que tiende al blanco, o se le acerca más que otro de su misma clase.'' (lit. ``Said about a colour or tone: with a tendency to white or closer to it than any other of the same class''). These basic meanings are straightforward and match this contextual sense.

\enumsentence{\label{ex:claro-nombre} Los otros nombres de modelos tenían un significado \textit{claro} (lit. ``The names of other models had a clear meaning'').}   
\enumsentence{\label{ex:claro-reina} La reina Sofía vestía un abrigo verde claro (lit. ``Queen Sofía was wearing a light green coat''). }

However, in (\ref{ex:claro-nombre}) it is harder to distinguish which is the contextual meaning according to the nuanced definitions provided in DRAE \citep{drae}:  ``Inteligible, fácil de comprender'' (lit. ``Intelligible, easy to understand''), ``Que se percibe o distingue bien'' (lit. ``Properly perceivable or
distinguishable''), ``Expresado sin reservas, francamente'' (lit. ``Expressed without reservations'').
Regardless the ambiguity of the contextual meaning, all these senses
are opposed to the basic sense and belong to different domains: \emph{claro} in (\ref{ex:claro-nombre}) alludes to LANGUAGE or COMMUNICATION, while the basic meaning is from the LIGHT or COLOR domain. Thus, we labeled the adjective as a metaphor in despite of not being able to determine exactly the contextual meaning.

\noindent \textbf{Pronominal Verbs}: Some verbs in Spanish present a pronominal form, which consists of a verb and a pronoun, either prepended and graphically separated from the verb form or as a clitic: \emph{se
arrepienten} (lit. ``they repent'') or as a clitic: \emph{no pueden arrepentirse} (lit. ``they cannot repent''). This pronoun can have multiple functions depending on its context of appearance, namely, reflexive, reciprocal\ldots ~Thus, it is important for annotators to be able to discern each use case. In our dataset pronouns are not within the scope of annotations but verbs are. This kind of lexical units is represented in CoMeta by three different tokens: a) verb and clitic pronoun: verb+se, e.g. \emph{olvidarse} (lit. ``to forget''); b) the verb form, e.g. \emph{olvidar} ; c) the pronoun.
In order to capture verbal metaphors and its semantic information,
we tagged options a) and b) in case of metaphorical expressions materialized through this structure. For instance, in example (\ref{ex:engancharse}), the presence of the clitic implies a difference in meaning. The pronominal variant of \emph{enganchar} (lit. ``to hook") in this context is used metaphorically, where the football player returns back to the league, so we tagged tokens \emph{engancharse} and \emph{enganchar}. 
\enumsentence{Garrido tendrá hoy un partido especial, sobre todo por si puede \emph{engancharse} a la
Europa League (lit. ``Garrido will have a special match today, mainly if he is able to rejoin the European League'').\label{ex:engancharse}}

\noindent \textbf{Multiword Expressions}: Multiword expressions, generally speaking, can be understood as the result of two or more words that co-occur with high frequency and act as a single lexical unit. MIPVU \citep{steenetal2010} prompts to annotate the contextual meaning of a MWE as a whole. However, in the actual annotation process, doubts arise as to whether some expressions can be considered a MWE or not. 

MIPVU used a list from the British National Corpus with MWEs as aid for their identification. In Spanish, there is no such resource, so we utilised the DRAE \citep{drae}. If an expression is registered in the dictionary with an individual entry, we treated it as a single lexical unit.  

MWEs included in dictionaries are often idiomatic, with non-compositional nor transparent meaning.  Since the overall meaning of an idiomatic expression rarely has anything to do with the sum of its constituents, they behave as a black box. In practice, \textit{corriente}  in example (\ref{ex:corriente-idiom}) is part of the idiom collected in DRAE \textit{estar al corriente}, which means ``to be aware or know about something''. Therefore it is not considered a lexical unit but a piece of a larger MWE which, in this case, is not metaphorical. 
On the contrary, \textit{corriente} (lit. ``current'') in (\ref{ex:corriente}) can be treated as a single lexical unit with a contextual meaning of ``trend'' or a group of people that share similar principles that opposes to its most basic sense that alludes to the movement of some fluids, \textit{corriente de aire} (lit. ``airflow''), it is annotated as a metaphor.

\enumsentence{Estaba al corriente de sus secretos. (lit. ``They were aware of their secrets''). \label{ex:corriente-idiom}} 
\enumsentence{Una \textit{corriente} cristiana que se originó en el siglo I. (lit. ``A christian current that was originated in the I century''). \label{ex:corriente}}

\begin{table*}[h]
\centering
\begin{tabular}{cccccc}
\cline{2-6}
 & \multicolumn{3}{c}{\textbf{VUAM}} & \multicolumn{2}{c}{\textbf{CoMeta}} \\ \cline{2-6} 
 & \textbf{Train} & \textbf{Dev} & \textbf{Test} & \textbf{Train} & \textbf{Test} \\ \hline
\multicolumn{1}{c}{\textbf{Metaphor}} & 8668 & 2372 & 3982 & 1713 & 432 \\ \hline
\multicolumn{1}{c}{\textbf{No\_Metaphor}} & 135896 & 34297 & 54347 & 91628 & 23342 \\ \hline
\multicolumn{1}{c}{\textbf{Total}} & \multicolumn{1}{l}{144564} & \multicolumn{1}{l}{36669} & \multicolumn{1}{l}{58329} & \multicolumn{1}{l}{93341} & \multicolumn{1}{l}{23774} \\ \hline
\end{tabular}
\caption{Number of metaphorical and non-metaphorical tokens in VUA and CoMeta datasets.}
\label{table:met-tokens}
\end{table*}

\subsubsection{Annotation Evaluation} \label{sec:agreement}

To analyse quantitatively the consistency of CoMeta annotations, we randomly selected the 10\% of sentences to be labeled by other annotators over the whole corpus. In other words, these sentences could belong either to train or test partitions. From this subset, 80\% of the sentences contained at least one metaphorical expression labeled as such by the main annotator of CoMeta. The purpose is to examine the consensus in the metaphorical annotations.

A total of 6 annotators participated in the evaluation and all of them were also Spanish native speakers with linguistic background knowledge. Each one reviewed 60 sentences randomly distributed and non-overlapping. As an aid for the task, we presented them the MIPVU guidelines and illustrative examples in advance. 
For each sentence, we extracted randomly 4 lexical units. We added a check-box next to each of these potential metaphorical expressions. Annotators must check those they deemed were holding metaphorical meaning in the context of that sentence. We included two additional options: one check-box to be marked in case there were no metaphorical expressions; and another one for annotators to write spotted metaphors that were not among the 4 candidates presented.
We computed inter-annotator agreement by means of Cohen's Kappa and obtained an average score of 0.631, which gives an account of the hardship and subjectivity of the task but also indicates a substantial consistency in the annotations.

\subsection{Data Analysis}

The most frequent metaphors arise from verbs, followed by nouns, adjectives and adverbs. Nevertheless, in political discourse texts, noun metaphors are more numerous than verbs, as shown in Table \ref{tab:data_analysis}.
Verbal metaphors usually involve verbs denoting motion or change of state,  e.g. \textit{abrir/cerrar} (lit. ``to open/close''), \textit{salir/entrar} (lit. ``to go in/out''), \textit{ascender/descender} (lit. ``to ascend/descend''), \textit{frenar/acelerar} (lit. ``to accelerate/brake''), \textit{partir/llegar} (lit. ``to leave/arrive''), and many others. Personifications are frequent as well (\ref{ex:personification}), through verbs that denote actions typically executed by an animate agent attributed to an inanimate entity (examples from CoMeta).

\enumsentence{Les \emph{atrapó} la miseria humana.  (lit. ``Human misery caught them''). \label{ex:personification}}

Adjectival metaphors arise in many cases through synesthesia and adjectives denoting physical dimensions applied to abstract or uncountable concepts (\ref{ex:oleaje}, \ref{ex:foto}).

\enumsentence{\emph{Tozudo} oleaje. (lit. ``Stubborn waves''). \label{ex:oleaje}}
\enumsentence{Foto \emph{rancia}. (lit. ``Rancid photograph''). \label{ex:foto}}

Regarding the domains of the conceptual mappings, we have observed several instances of metaphorical expressions that depict politics in terms of the construction field (\ref{ex:construir}, \ref{ex:solida}), and a virus or a disease as war (\ref{ex:vencer}, \ref{ex:coronavirus}).

\enumsentence{\label{ex:construir}Es imposible \emph{construir} un proyecto de Estado. (lit. ``It is impossible to build a State project''). }
\enumsentence{\label{ex:solida} La candidatura de Osaka es muy \emph{sólida} (lit. ``Osaka’s candidacy is very solid''). }
\enumsentence{\label{ex:vencer} Unidos conseguiremos de nuevo \emph{vencer} al virus (lit. ``Together we will defeat the virus again''). }
\enumsentence{\label{ex:coronavirus} El único \emph{arma} terapéutica que tenemos en este momento para \emph{luchar} contra el coronavirus (lit. ``The only therapeutic weapon available at this time to fight
against coronavirus''). }

\begin{table*}[ht]
\centering
\begin{tabular}{l|l|lll}
\hline
\textbf{Dataset} & \textbf{Model}    & \textbf{Prec}  & \textbf{Rec}   & \textbf{F1}    \\ \hline
                 & ixabertes\_v1     & 71.99          & 59.49          & 65.15          \\
CoMeta           & mdeberta-v3-base  & \textbf{78.70}          & 59.03          & \textbf{67.46}          \\
                 & xlm-roberta-large & 75.57          & \textbf{60.88}          & 67.44 \\ \hline \hline
                 & deberta-large     & \textbf{79.95} & 68.50          & \textbf{73.79} \\
VUAM             & deberta-base      & 73.06          & \textbf{73.07} & 73.07          \\
                 & xlm-roberta-large & 77.99          & 68.00          & 72.65 \\ \hline
VUAM SOTA & MelBERT & 76.4 & 68.6 & 72.3 \\ \hline
\end{tabular}
\caption{Monolingual results for Spanish and English.}
\label{tab:mono}
\end{table*}

% \usepackage{multirow}
\begin{table*}[h]
\centering
\begin{tabular}{c|c|c|ccc}
\hline
\textbf{Train Dataset} & \textbf{Test Dataset} & \textbf{Model} & \textbf{Prec} & \textbf{Rec} & \textbf{F1} \\ \hline
CoMeta                                    & VUAM                                 & mdeberta-v3-base                & 76.28                          & 58.8                          & 66.41                        \\ 
VUAM                              & CoMeta                                   & xlm-roberta-large               & 73.95                          & 70.36                         & 72.11                        \\ \hline
\end{tabular}
\caption{Results from cross-lingual experiments after 4 epochs.}
\label{tab:cross}
\end{table*}

\section{Evaluation}

In this section we present the experiments on metaphor detection in Spanish and English. Furthermore, we also report the results of the first supervised cross-lingual experiments for metaphor detection. The main objective of the cross-lingual evaluation setting was to examine which kind of metaphors carried more often across languages.

\subsection{Datasets}

The two datasets used for experimentation are the VUAM dataset \citep{steenetal2010} in English, and CoMeta in Spanish. With respect to the VUAM dataset \citep{steenetal2010}, we employed the original train and test splits provided in the shared task \citep{leong-etal-2020-report}. We also extracted a development set by splitting the training set (0.8-0.2). Using the original train and test partitions will allow us to compare with previous results. In the case of CoMeta, and due to its smaller size, we did not create a development split. Table \ref{table:met-tokens} provides the stats for each corpus. It should be noted that both datasets are imbalanced. In the case of CoMeta we decided not to alter this distribution since it represents the frequency of metaphor in natural language texts.

\subsection{Experimental Setup}

We perform experiments in two evaluation settings: monolingual and cross-lingual. For the monolingual setting, we evaluate on the English and Spanish datasets using the most commonly used large language models for each of the languages. In the cross-lingual setting we evaluate the best performing 
multilingual language model for each language in a zero-shot scenario, namely, fine-tuning in a source language and making the predictions in another language, not seen during fine-tuning.

\noindent \textbf{Monolingual Experiments:} The experiments performed in this setting aimed to establish a baseline with respect to the state-of-the-art in metaphor detection for English using the VUAM corpus, currently represented by MelBERT \citep{melbert}. This baseline will also help us to judge the performance on the CoMeta dataset. We picked the 9 most commonly used large language models for each language, both in their \emph{base} and \emph{large} versions (DeBERTa also includes mDeBERTa, a multilingual base model pre-trained for 100 languages). For English we experimented with BERT \citep{devlin2018bert}, RoBERTa \cite{liu2019roberta}, DeBERTa \citep{He2021DeBERTaV3ID} and XLM-RoBERTa \citep{xmlroberta}. 

With respect to Spanish, we used BETO \citep{CaneteCFP2020}, ixabertes\_v1 and ixabertes\_v2\footnote{Avalaible in \url{http://www.deeptext.eus/es/node/2}}, ixambert \citep{otegi2020conversational}, RoBERTa-BNE models \citep{gutierrezfandino2022} and the multilingual models mDeBERTa and XLM-RoBERTa (base and large). Every model was fine-tuned via the Huggingface Transformers library \citep{wolf-etal-2020-transformers}.

We performed hyperparameter tuning for batch size (8, 16, 36), linear decay (0.1, 0.01), learning rate (in the [1e-5-5e-5] interval) and epochs from 4 to 10. We keep a fixed seed of 42 for experimental reproducibility and a sequence length of 128. A warm-up of 6\% is specified. The results of the hyperparameter tuning showed that after 4 epochs development loss started to increase, so every result reported here is obtained by performing 4 epochs only. Furthermore, the results of the best models are chosen according to their performance on the development for each language. Finally, due to presentation reasons, we decided to include only the best three models: the best \textit{base} and \textit{large} models for each language, the best Spanish monolingual and the best multilingual for English. These are the results included in Table \ref{tab:mono}. Results of all models are gathered in Appendix \ref{sec:appendix}, Table \ref{tab:mono-all}.

\noindent \textbf{Cross-lingual Experiments:} The aim of this experiments is to explore: a) whether a model trained with metaphorical annotations from one language can achieve good results when evaluating metaphors in another language and, b) to what extent metaphors are shared between these languages. Thus, in this setting we picked the best performing multilingual model for each of the two monolingual evaluations and apply them in a zero-shot cross-lingual manner, namely, by fine-tuning the language model on the English dataset and evaluating it with the Spanish one, and viceversa, using the best hyperparameter configuration obtained in the monolingual setting.

\begin{table*}[ht]
\centering
\begin{tabular}{cccccccc}
\multicolumn{1}{l}{}                                 & \multicolumn{1}{l}{}    & \multicolumn{1}{l}{\textbf{Monolingual}}                                                             & \multicolumn{1}{l}{\textbf{Cross-lingual}}                                                         & \multicolumn{1}{l}{}                 & \multicolumn{1}{l}{}    & \multicolumn{1}{l}{\textbf{Monolingual}}                                                                    & \multicolumn{1}{l}{\textbf{Cross-lingual}}                                                     \\ \hline
\multicolumn{1}{c|}{\multirow{9}{*}{\textbf{VUAMC}}} & \multicolumn{1}{c|}{FP} & \multicolumn{1}{c|}{\begin{tabular}[c]{@{}c@{}}get 33\\ got 22\\ little 16\\ go 16\end{tabular}}     & \multicolumn{1}{c|}{\begin{tabular}[c]{@{}c@{}}get 21\\ got 20\\ go 14\\ bloody 12\end{tabular}}  & \multicolumn{1}{c|}{}                & \multicolumn{1}{c|}{FP} & \multicolumn{1}{c|}{\begin{tabular}[c]{@{}c@{}}crecimiento 3\\ paso 3\\ espacio 3\\ repaso 2\end{tabular}}  & \begin{tabular}[c]{@{}c@{}}contempla 4\\ crecimiento 3\\ espacio 3\\ repaso 2\end{tabular}    \\ \cline{2-4} \cline{6-8} 
\multicolumn{1}{c|}{}                                & \multicolumn{1}{c|}{FN} & \multicolumn{1}{c|}{\begin{tabular}[c]{@{}c@{}}got 13\\ away 12\\ back 12\\ subject 10\end{tabular}} & \multicolumn{1}{c|}{\begin{tabular}[c]{@{}c@{}}back 14\\ got 12\\ plant 12\\ get 11\end{tabular}} & \multicolumn{1}{c|}{\textbf{CoMeta}} & \multicolumn{1}{c|}{FN} & \multicolumn{1}{c|}{\begin{tabular}[c]{@{}c@{}}estabilidad 6\\ gran 4\\ ocupa 4\\ dimensión 4\end{tabular}} & \begin{tabular}[c]{@{}c@{}}estabilidad 6\\ ocupa 4\\ dimensión 4\\ seguimiento 3\end{tabular} \\ \cline{2-4} \cline{6-8} 
\multicolumn{1}{c|}{}                                & \multicolumn{1}{c|}{TP} & \multicolumn{1}{c|}{\begin{tabular}[c]{@{}c@{}}make 50\\ take 33\\ way 32\\ got 26\end{tabular}}     & \multicolumn{1}{c|}{\begin{tabular}[c]{@{}c@{}}make 48\\ take 34\\ way 33\\ got 27\end{tabular}}  & \multicolumn{1}{c|}{}                & \multicolumn{1}{c|}{TP} & \multicolumn{1}{c|}{\begin{tabular}[c]{@{}c@{}}marco 8\\ ola 6\\ abrir 4\\ escenario 4\end{tabular}}        & \begin{tabular}[c]{@{}c@{}}marco 8\\ ola 6\\ abrir 4\\ escenario	4\end{tabular}               \\ \hline
\end{tabular}
\caption{
Top-4 terms of false positive (FP), true positive (TP) and false negative (FN) predictions from experiments performed with VUAM and CoMeta in monolingual and cross-lingual scenarios.
}
\label{tab:error_analysis}
\end{table*}
\subsection{Results}

The first interesting result of our experiments is that the general purpose DeBERTa-large language model performs slightly better than the metaphor-specific MelBERT, with the base version not far behind. With respect to Spanish, the results are not as high in general as those obtained for English. In particular, the performance of XLM-RoBERTa-large for Spanish is substantially lower than for English. Apart from many other factors that may be involved, we attribute these lower results to the smaller size of the Spanish training set. It is also interesting to note that a base multilingual model, mDeBERTa, is the best performing model for Spanish, obtaining very similar results to XLM-RoBERTa-large. Still, the low results of the state-of-the-art models show that this remains a highly difficult task.

For the cross-lingual results, we picked the best multilingual model for each of the monolingual settings, mDeBERTa for Spanish and XLM-RoBERTa-large for English. The results reported in Table \ref{tab:cross} show that the zero-shot performance is remarkably high, which is quite surprising, especially if we consider the performance of XLM-RoBERTa-large for Spanish. In fact, these results show that XLM-RoBERTa obtains better results for Spanish when fine-tuned in English. Next section will provide some analysis to attempt to explain this phenomenon. In any case, the results obtained for Spanish are promising and encourage us to continue improving the annotated resources for this language.

\subsection{Error Analysis}

In Table \ref{tab:error_analysis} we enumerated the most frequent predictions which are potentially interesting for error analysis. These predictions correspond to the model with best performance, DeBERTa-large in the case of VUAM and mDeBERTa for CoMeta. False positives (FP) represent lexical units that were labeled wrongly as metaphorical. False negatives (FN) include metaphorical expressions that were not detected as such by the model, whereas true positives (TP) gather which metaphorical expressions were accurately identified.

The FP and FN from the monolingual setup of VUAM show mostly verbs that tend to form collocations, like \emph{go} or \emph{get}, or highly lexicalised terms, such as \emph{little, away, subject} or \emph{back}. The high occurrence of these lexical units both with metaphorical and literal meaning and the high degree of polysemy difficult the possibility to learn patterns. In the case of CoMeta, FP and FN comprise terms that scarcely appear in our dataset with metaphorical meaning or in similar proportions with metaphorical and literal tags.

With respect to TP, in VUAM predictions, we can find again terms that occur in the dataset very frequently conforming collocations and phrasal verbs, which are commonly tagged as metaphors. Right predictions in CoMeta present lexical units that only appear with metaphorical meaning, such as \emph{ola} (lit. ``wave'') in relation to the virus domain, which does not occur in CoMeta with a literal sense.

Results from cross-lingual experiments show an outcome which resembles that of the monolingual setup. This similarity between
both setups was noticeable from the scores of the evaluation metrics in Tables \ref{tab:mono} and \ref{tab:cross}. This suggests that, due to its current size, training on CoMeta obtains worse results than training in English. We hypothesize that, in addition to the size, the high frequency of commonly used verbal lexical units that are labelled as metaphors in both datasets help to obtain such good results in the cross-lingual setting.

% \usepackage{multirow}
% \usepackage{multirow}
% \usepackage{multirow}

\section{Conclusions and Future Work}

In this work we have created CoMeta, which to the best of our knowledge is the largest dataset with metaphor annotations in Spanish composed of texts from various domains to be publicly available. We also discussed in detail the main issues that emerged during the annotation process for Spanish. In order to evaluate the quality of CoMeta's annotations we carried out a series of experiments in both monolingual and cross-lingual environments, using the largest dataset with metaphor annotations in English, the VUAM corpus, as reference point.. Moreover, we set a new state of the art on the task of metaphor detection in English and set a strong baseline for the task in Spanish, which hopefully will encourage researchers to continue with this line of work.

The aim of this work is to lay the foundations for future development on metaphor detection in Spanish and cross-lingually. Regarding the dataset, a future line of work would introduce more fine-grained tags that represent the different kinds of metaphorical expressions. This task should be performed by multiple annotators, in order to explore agreement over the whole dataset, as well as to observe if doubtful cases share any feature that could be leveraged for their identification. The presence of more fine-grained tags would also enable a deeper statistical analysis of CoMeta that could be exploited to study how metaphor manifests in Spanish and whether there are similarities with the usage of metaphor in other languages. 

Results obtained from our experiments encourage future research to continue with cross-lingual approaches. We hypothesize that these results may be due to the difference in size of the training data in both languages or the application of MIPVU guidelines to Spanish, which is not the language it was originally designed for. Future experimental work is needed to test these interpretations, which could benefit from the extension of the annotations in CoMeta we just mentioned.

\section*{Acknowledgements}
This work has been supported by the HiTZ center and the Basque Government (Research group funding IT-1805-22). Elisa Sanchez-Bayona is funded by a UPV/EHU grant ``Formación de Personal Investigador''. Rodrigo Agerri acknowledges the support from the RYC-2017–23647 fellowship (MCIN/AEI /10.13039/501100011033 y por El FSE invierte en tu futuro), and from the projects DeepKnowledge (PID2021-127777OB-C21) by MCIN/AEI/10.13039/501100011033 and FEDER Una manera de hacer Europa, and Disargue (TED2021-130810B-C21) by MCIN/AEI /10.13039/501100011033 and European Union NextGenerationEU/PRTR .

\section*{Limitations}

The presented dataset is limited in size compared to its English counterpart, the VUAM corpus. Therefore, a second version of CoMeta augmented with more texts of domains where metaphors are more abundant should be a priority of future work. This would be important both for monolingual and cross-lingual results, especially to analyze the cross-lingual transfer behaviour of the multilingual models. Furthermore, the process of metaphor labelling is inherently subjective and annotator-dependent, since personal experience and socio-cultural features may influence the identification of metaphors, as well as the domain of collected texts. Thus, the incorporation of a variety of annotators would alleviate this issue. 
In any case, we believe that CoMeta represents a worthy first contribution towards multilingual and cross-lingual metaphor detection and that the results obtained in this paper can be improved by further developing CoMeta to be a dataset of size similar to VUAM. Finally, even if we reported state-of-the-art results, the overall low performance means that further work on this task is required.

\bibliography{anthology,custom}
\bibliographystyle{acl_natbib}

\appendix

\section{Appendix}\label{sec:appendix}

In Table \ref{tab:mono-all} we gather the performance of all models used in monolingual experiments over the test set. For each model, we only included the version that achieved the highest F1 score with the specified parameters after 4 epochs. Bold results correspond to the model that obtained top performance, while underscored results correspond to the second best score.

\begin{table*}[ht]
\centering
\begin{tabular}{llrrrlcc}
\hline
\textbf{Dataset} & \textbf{Model} & \textbf{Batch Size}  & \textbf{Weight Decay}  & \textbf{Learning Rate}  & \textbf{F1}    \\ \hline \hline
               \multirow{14}{*}{CoMeta}
                & \multicolumn{5}{c}{\emph{Monolingual}} \\ \cline{2-6}
               & bertin  & 8 & 0.01 & 0.00003 & 61.56 \\
                & beto  & 8 & 0.01 & 0.00005 & 64.28 \\
                 & electricidad  & 8 & 0.1 & 0.00005 & 61.18  \\
                  & \underline{ixabertes\_v1}  & 8 & 0.01 & 0.00005 & \underline{65.15}          \\
                 & ixabertes\_v2 & 8 & 0.01 & 0.00005 & 64.79          \\
                  & ixambert    & 8 & 0.1 & 0.00005 & 62.04         \\
                  & roberta-large-bne    & 16 & 0.1 & 0.00001 & 62.02          \\
                  & roberta-base-bne  & 8 & 0.1 & 0.00005 & 63.07 \\ \cline{2-6}
                  & \multicolumn{5}{c}{\emph{Multilingual}} \\ \cline{2-6}
                    & mbert    & 8 & 0.01 & 0.00005 & 61.78        \\
                  & \textbf{mdeberta-v3-base} & 8 & 0.01 & 0.00005  & \textbf{67.46}  \\
                  & xlm-roberta-base & 8 & 0.1 & 0.00005 & 63.82  \\
                  & xlm-roberta-large & 8 & 0.01 & 0.00002 & 67.44          \\
                \hline \hline
               \multirow{11}{*}{VUAM} 
               & \multicolumn{5}{c}{\emph{Monolingual}} \\ \cline{2-6}
                 & bert-base & 16 & 0.01 & 0.00005 & 69.99 \\
                 & bert-large & 32 & 0.01 & 0.00005  & 71.67 \\
               & deberta-base & 32 & 0.1 & 0.00005 & 73.07 \\
               & \textbf{deberta-large} & 8 & 0.01 & 0.00002 &  \textbf{73.79} \\
                & roberta-base & 8 & 0.01 & 0.00005 & 70.11          \\
                & roberta-large & 32 & 0.1 & 0.00005 & 72.69          \\ \cline{2-6}
                & \multicolumn{5}{c}{\emph{Multilingual}} \\ \cline{2-6}
                & mdeberta-v3-base & 16 & 0.01 &0.00005 & 70.40          \\
                & xlm-roberta-base & 8 & 0.1 & 0.00002 & 66.59 \\
                 & \underline{xlm-roberta-large} & 32 & 0.1 & 0.00003 & \underline{72.11} \\ \hline
\end{tabular}
\caption{Results from monolingual experiments with all models, trained over 4 epochs, for English and Spanish.}
\label{tab:mono-all}
\end{table*}

\end{document}